\documentclass[USenglish]{article} 

\usepackage[utf8]{inputenc} 
\usepackage{lmodern}
\usepackage{microtype}
\usepackage[numbers,square,sort&compress]{natbib}

\usepackage{amsthm} 
\usepackage{amsmath} 
\usepackage{amssymb}
\usepackage{graphicx} 
 \usepackage{url}

\theoremstyle{plain}

\theoremstyle{definition}

\usepackage{authblk}

\title{Automatic Differentiation with Constraints in Forward and Inverse Problems}

\author[1]{Nikolay Yavich\thanks{Corresponding author, \texttt{n.yavich@skoltech.ru}}}
\author[1]{Alexander Ryabov}
\author[1]{Viacheslav Naumov}
\author[2]{Sayan Ranu}
\author[2]{N. M. Anoop Krishnan}
\author[1,3]{Evgeny Burnaev}
\author[1]{Vladimir Vanovskiy}

\affil[1]{Artificial Intelligence Center, Skolkovo Institute of Science and Technology, 121205 Moscow, Russia}
\affil[2]{Yardi School of Artificial Intelligence, Indian Institute of Technology Delhi, Hauz Khas, New Delhi, India 110016}
\affil[3]{Learnable Intelligence Group, Autonomous Non-Profit Organization Artificial Intelligence Research Institute (AIRI), 105064 Moscow, Russia}

\date{} 

\begin{document}

\maketitle

\begin{abstract}
Approximate solutions of partial differential equations (PDEs) on unstructured grids can be represented on a graph. Numerical optimization involving stiff evolutionary and static PDEs often requires solving a large system of equations, which can be treated as a constraint. In this work, we present a fully differentiable graph-based implicit solver that incorporates this constraint. We evaluate its performance across four key tasks: a) linear and nonlinear forward modeling of evolutionary and static problems, b) solving inverse problems to estimate equation coefficients, c) solving inverse problems to estimate the right-hand side, and d) learnable graph coarsening to accelerate forward modeling. The stability and differentiability of our solver enable significant reductions in the complexity of forward modeling while efficiently addressing related inverse problems. These capabilities make it a powerful tool for different physics and engineering applications, in particular for geophysics, where inverse problems are commonly solved in order to reduce the model uncertainty.

\vspace{0.5em}
\end{abstract}

\section{Introduction}

In recent years, graph neural networks have emerged as a powerful tool for handling data on unstructured grids \citep{gcn, gat, graphsage, gin}. Using message-passing mechanisms, GNNs are highly effective in capturing local dependencies and updating node features through their connections. This capability has sparked increasing interest in applying GNNs to PDE solvers, particularly in problems where the data can be represented as a graph, such as Voronoi grids and finite-volume discretizations.

The finite-volume (FV) method is a well-established numerical technique for solving PDEs, especially on unstructured grids, as it ensures local conservation properties. Integrating FV methods with GNN-based solvers provides a structured, grid-free representation of the domain, while preserving a rigorous framework for solving governing equations. In this context, the development of differentiable solvers is an especially promising research direction.

One key advancement explored in this work is the application of automatic differentiation (AD) \cite{Naumann} within GNN-based solvers, particularly for tasks such as graph coarsening and solving inverse problems. AD has proven to be an invaluable tool for efficient gradient-based optimization.
The use of AD in inverse problems involving hyperbolic equations was demonstrated in \cite{ZhuDarve2021}.

In our work, we focus on parabolic partial differential equations, which are crucial in various applications, including engineering and the oil and gas industry. However, the high stiffness of the resulting evolutionary problems necessitates the use of implicit time-stepping schemes due to their stability properties. Additionally, we address elliptic partial differential equations, where a distinctive challenge lies in solving large and sparse linear systems.

Automatic differentiation (AD), \cite{AD_2000, AD_Review},
is an established tool to form gradients of functions. 
Although some AD libraries support implicit relations between variables, e.g., \cite{JAX}, their capabilities for handling sparse linear solvers are currently quite limited.
The JAX \citep{JAX_spsolve} solver is based on the QR matrix factorization, which is suitable for small and moderate size problems, but for larger problems it will be too slow and memory-consuming.  
Analogous Torch functionality \citep{PyTorch_Dense_Solve, PyTorch_Sparse} is only being developed.
In our work, we show how 
a differentiable implicit solver can be built on the basis of any automatic differentiation framework.
To address this, we developed custom autograd functions tailored to our needs.

Building on this foundation, we propose an integrated approach that combines GNNs, finite-volume methods, implicit time-stepping, and automatic differentiation to create a fully differentiable pipeline for both forward and inverse modeling. Our approach ensures differentiability with respect to various components, including grid cell locations, PDE coefficients, and the right-hand side, providing a flexible and robust framework for tackling both forward modeling and inverse problems.

We should emphasize that our focus lies 
in design and use of asymptotically correct (non-trainable) differentiable implicit FV solvers 
rather than in training surrogate models for them
(see Section \ref{ref:rel_work} for more details).

Specifically, we focus on the following key problems:

\begin{itemize} 
\item Forward modeling of stiff evolutionary and static problems; 
\item Solving inverse problems to estimate equation coefficients; 
\item Solving inverse problems to estimate unknown right-hand side terms; 
\item Graph coarsening to accelerate forward modeling. 
\end{itemize}

We demonstrate the capability of the proposed approach to effectively address all these problems, showcasing its flexibility and ease of implementation for geophysical modelling and other PDE-based applications.
\section{Related Work}
\label{ref:rel_work}

A significant number of recent works can be characterized as surrogate modeling. In the context of forward problems, the goal is to approximate the Green's function of the equation, as seen in \cite{alet2019,Lu2021DeepONet,li2021FNO}.
The work of \cite{Nastorg2024} involved implicit Graph Neural Networks (GNNs) for this task. 
Implicit deep learning, \cite{ElGhaoui2021}, also referred to as deep equilibrium models (DEQ),
substitutes conventional activation functions at particular layers with a nonlinear equilibrium equation completed with a prediction equation.
DEQ combined with GNNs are known as implicit GNNs, \cite{gu2021}. Their benefit
is that data propagates instantly within the graph, 
rather than with finite speed as in the case of conventional deep
learning.
The work of \cite{horie2024} proposed a learnable approach for computing fluxes in the FVM, motivated by convection-dominated problems.

For inverse problems, the aim is to approximate the inverse of a potentially nonlinear forward operator using a given dataset. For instance, \cite{qzhao2022} employed GNNs for this problem. In \cite{jessica2023}, the authors enhanced GNNs with novel geometric features to improve prediction accuracy. 

The main drawback of these approaches is the massive data requirements for training (ranging from gigabytes to terabytes).  Also, the training is computationally expensive, often requiring hundreds to thousands of GPU hours for training. More importantly, the obtained solution is less interpretable compared to solvers employing the classical numerical schemes.
However, as will be shown below, our method outperforms PINNs in solving inverse problems both in terms of quality and computational expenses.

In traditional methods for inverse problems, as described in \cite{zhdanov2002}, optimization is performed for each new dataset. These algorithms rely on the derivation of Jacobians, a tedious and error-prone task. The availability of automatic differentiation (AD) pipelines can help avoid this issue. For example, \cite{ZhuDarve2021} presented an AD-based acoustic data inversion method combined with explicit Euler time discretization for the wave equation. In our work, we follow a similar direction by demonstrating how implicit schemes can be integrated into an AD pipeline and further combined with GNNs.

Our approach is closely related to physics-informed neural networks (PINNs) \cite{PINN_2019,kharazmi2019VPINNs}, \cite{PINN_2021}, and their extensions, 
such as graph convolutional PINNs \cite{Gao2022}, as both methods can be considered self-supervised learning techniques.


\section{Methodology}
\label{sec:thory}

\subsection{Governing equations}

In this work, we mainly consider a parabolic partial-differential equation in some domain $V$,
\begin{equation}
    \displaystyle\frac{\partial u}{\partial t} - \textrm{div}(K \nabla u) = f,   \quad 0<t<T,  \quad (x,y)\in V,
    \label{eq:parab}
\end{equation}
where $u(x,y,t)$ is the unknown variable, $K(x,y)$ is a positive coefficient, $f(x,y,t)$ is the source term.  The above equation is completed with the initial condition $u = u^0(x,y)$ at $t=0$
and zero Neumann boundary conditions on the domain boundary. 

Assume that we are given with a set of points $C_s$, $1\leq s \leq N$. We use them to generate a Voronoi grid, \eqref{fig:teaser}. A numerical solution to \eqref{eq:parab} can be received by combining the finite volume method (FVM) for spatial discretization \citep{Herbin2000, Kuznetsov2007}
and the implicit Euler scheme
for temporal discretization,
\begin{equation}
    D \displaystyle\frac{u^{k+1}-u^k}{\tau} + A \cdot u^{k+1} = D  f^{k+1}, \quad 0\leq k \leq n-1. 
    \label{eq:implicit_scheme}
\end{equation}
where $u^k \in \mathbb{R}^N$, $0\leq k \leq n$
is the numerical solution at the respective time step, $A$ is the finite volume matrix, $D$ is
the diagonal matrix of Voronoi cells areas, and $f^{k}$ is the source vector. 

We implemented the finite volume method within
the graph neural network (GNN) framework
with the message passing paradigm. 
Our GNN solver uses weights for the edges of the graph,
\begin{equation}
w_{ij} = -A_{ij}.
\end{equation}
Since every non-zero off-diagonal matrix entry corresponds
to some edge of the graph,
the GNN message massing mechanism could be used to implement matrix-vector multiplication, $A \cdot u^k$.

\subsection{Problem Setup}

Optimization in both static and evolutionary forward and inverse problems can be written in the following form,
\begin{equation}
L(\theta) = \sum_{k=1}^n \|R \cdot u^k(\theta) - u^k_{obs}\|^2
+ S(\theta),
\label{eq:gen_optim}
\end{equation}
where 
$\theta\in\mathbb{R}^M$ is an unknown parameter requiring estimation, 
$n$ is the number of temporal measurements (1 for static problems), 
$u^k(\theta) \in \mathbb{R}^N$ is the modelled variable,
$R:\mathbb{R}^N \to \mathbb{R}^m$ is the measurement operator,
$u^k_{obs}\in\mathbb{R}^m$ is measured data at the temporal point $k$, and $S(\theta)$ is a stabilizer.
The role of operator $R$ is to simply keep in a given vector 
several entries (wherever sensors are located) while removing the other entries.

Both static problems and evolutionary problems involving implicit time-stepping will require solution of a large sparse equation system,
\begin{equation}
A(\theta) \cdot u^k(\theta) = f^k (\theta), \quad 1\leq k\leq n,
\label{eq:system_k}
\end{equation}
forming a constraint to \eqref{eq:gen_optim}.

A naive approach might be to eliminate $u^k(\theta)$ and receive an unconstrained optimization problem,
\begin{equation}
L(\theta) =  \sum_{k=1}^n 
\|R \cdot A(\theta)^{-1} \cdot f^k(\theta) - u^k_{obs}\|^2
+ S(\theta),
\end{equation}
However, this is not feasible since $A(\theta)^{-1}$ is a large dense matrix not suitable for manipulations including backward propagation. 

An alternative approach might involve solving the implicit equation through optimization techniques commonly used in deep learning, such as constructing a minimization problem for equation \eqref{eq:system_k} combined with a data loss function, potentially avoiding matrix inversion. 
However, matrices arising from discretization of elliptic PDEs, like \eqref{eq:system_k}, are highly ill-conditioned. Incorporation of such a matrix into optimization problems would lead to very slow convergence of the minimization algorithms (both stochastic and deterministic). 
Practically, \eqref{eq:system_k} is solved with either a direct or preconditioned iterative solver.

Let us notice the following. Firstly, automatic differentiation systems that are typically used for minimization of \eqref{eq:gen_optim}, may not
process implicit relations like \eqref{eq:system_k}
or the respective functionality may be limited.
Secondly, custom developed fast parallel solvers are commonly linked
to solve \eqref{eq:system_k} in practical modelling.
To address both of these cases, we will derive formulae for
forward and backward propagations in the next subsection.



\subsection{Forward and Backward Propagation}

 We will assume that the system matrix in \eqref{eq:system_k} is symmetric and positive-definite, what is typically the case within implicit time-stepping and static problems.
Let us rewrite the system as
\begin{equation}
A\cdot u = b \quad \text{or} \quad \sum_j a_{ij} u_j = b_i,  
\label{eq:system}
\end{equation}
for brevity. The matrix $A$ is large and sparse thus special formats are used to store its non-zero values only, e.g. compressed sparse row (CSR) or coordinate (COO) formats.

During the forward propagation stage, we have to solve \eqref{eq:system},
what could be formally written as, 
\begin{equation}
u = A^{-1} \cdot b.
\label{eq:sol}
\end{equation}
Practical algorithms avoid storing $A^{-1}$ and multiplication by it. 

During the backward propagation stage, we have to compute the gradients of $u$ with respect to $A$ and $b$.
The gradient with respect to $b$ could be found as follows, 
\begin{equation}
\quad \sum_j a_{ij} \frac{\partial u_j}{\partial b_k} = \delta_{ik}, 
\end{equation}
thus, 
\begin{equation}
\nabla_b u = A^{-1}.
\label{eq:grad_b}
\end{equation}
The gradient with respect to $A$ has the following form,
\begin{equation}
\sum_j \frac{\partial a_{ij}}{\partial a_{km}} u_j + \sum_j a_{ij} \frac{\partial u_j}{\partial a_{km}} = 0, 
\end{equation}
thus,
\begin{equation}
\nabla_A u = -A^{-1} \cdot P \cdot u.
\label{eq:grad_a}
\end{equation}
where $P$ is a perturbation matrix, i.e. a matrix of zeros with just a single entry of 1.

Now, these expressions can used to compute 
 the gradient of the loss \( L \) with respect to $A$ and \( b \)  efficiently. Notice
\begin{equation}
\nabla_b L = \nabla_u L \cdot \nabla_b u, 
\quad \text{and} \quad 
\nabla_A L = \nabla_u L \cdot \nabla_A u,
\label{eq:grad_L_chain}
\end{equation}
where $\nabla_u L$ is the gradient of the loss with respect to the solution \( u \), propagated back from subsequent stages in the computational graph. 

Substituting \eqref{eq:grad_b} and \eqref{eq:grad_a}
into \eqref{eq:grad_L_chain}, we receive,
\begin{equation}
\nabla_b L = \nabla_u L \cdot A^{-1}, 
\label{eq:grad_L_chain_2}
\end{equation}
and
\begin{equation}
\nabla_A L = -\nabla_u L \cdot A^{-1} \cdot P \cdot u = -\nabla_b L \cdot P \cdot u.
\label{eq:grad_L_chain_3}
\end{equation}
The last expressions imply the following computational algorithm for the backward propagation stage: 
\begin{enumerate}
    \item  Compute $\nabla_b L$ with \eqref{eq:grad_L_chain_2}, which is equivalent to solving of a single linear system.
    \item  Compute $\nabla_A L$  by multiplying the respective entries of $\nabla_b L$  and $-u$ by each other. 
\end{enumerate}
Thus computational complexity of the backward propagation stage is equivalent to the complexity solution of a single equation system.
Since $A$ is stored in a sparse matrix format,
\eqref{eq:grad_L_chain_3} is reduced to differentiating only non-zero entries.

We should emphasize there is a close conceptual relationship between forward and backward propagation in automatic differentiation (AD) and the notions of direct and adjoint states in constrained optimization. In forward mode AD, sensitivities of outputs with respect to inputs are propagated along the computational graph in the same direction as the evaluation of the primal function, mirroring the direct approach in optimization where the state equations are solved forward in time to obtain solution and its associated sensitivities. 
Conversely, backward (or reverse) mode AD propagates derivatives from outputs to inputs by traversing the computational graph in reverse, which is equivalent to the adjoint method in optimization: instead of directly computing sensitivities by forward perturbation, one introduces adjoint variables governed by backward equations, enabling efficient computation of gradients with respect to a large number of parameters. Thus, the duality between forward and backward modes in AD reflects the duality between direct and adjoint formulations in constrained optimization, with both frameworks exploiting structural properties of the underlying equations to reduce computational cost in large-scale problems,
\cite{Backpropagation,griewank2008evaluating}.


\section{Experiments}
\label{headings}

\subsection{Implementation}

\begin{figure*}[htb]
    \centering
    \includegraphics[width=0.75\textwidth]{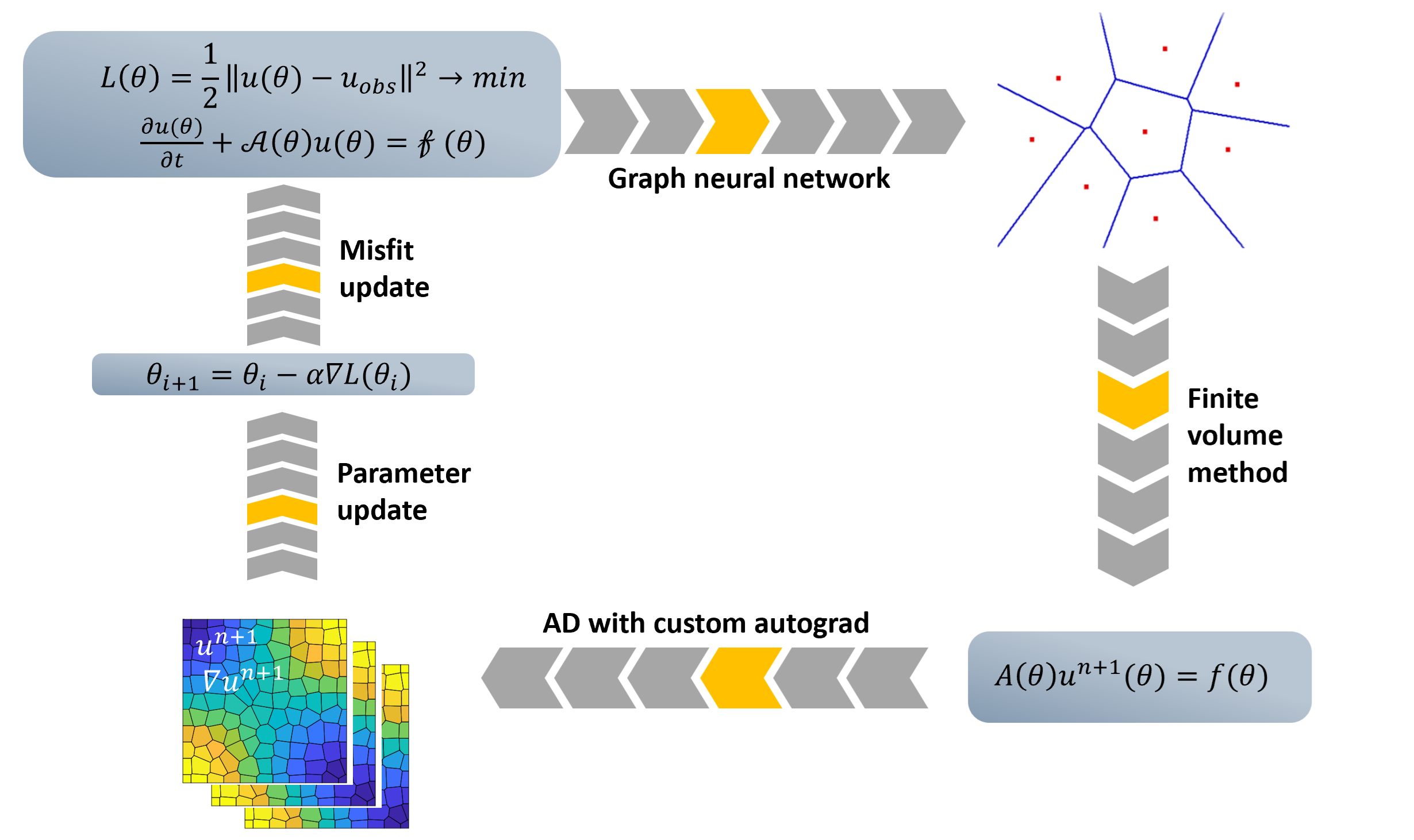}
    \caption{Differentiable modeling pipeline 
     developed in this work combines graph neural networks, the finite-volume method, implicit time-stepping and can be applied to optimization in forward and inverse problems. }
    \label{fig:teaser}
\end{figure*}

Our differentiable graph simulator is implemented based on the message passing paradigm in a graph neural networks. For implementing such a solver, we used Pytorch Geometric \cite{fey2019} framework.
Formalas \eqref{eq:grad_L_chain_2} and \eqref{eq:grad_L_chain_3} were implemented
in a PyTorch custom autograd function.
Optimization of graph coarsening is implemented with the method for autodifferentiation of
the 2D Voronoi tessellation, \cite{diffvoronoi}.
A schematic illustration of our pipeline is given in
Fig.~\ref{fig:teaser}.

\subsection{Learnable Graph Coarsening for the Linear Parabolic Equation}
\label{ss:fowrd}

Let a polygonal domain be filled with heterogeneous porous media. We assume that a slightly compressible fluid is injected at one point and pressure is measured at some other points.
With minor simplifications \citep{Chen2007} the fluid flow
is described by the parabolic equation,
\eqref{eq:parab}.

In these experiments, the coarsening process is guided by a graph neural network (GNN),  
which learns soft cluster assignments to map nodes from the fine grid to coarser clusters.  
Our approach builds upon \cite{shumilinself}, but instead of solving a global optimization problem for point locations,  
we adopt a graph coarsening method.  
Additionally, in this work, we replace the explicit time-stepping used in the baseline \cite{shumilinself}  
with the implicit time-stepping approach, \eqref{eq:implicit_scheme}.  

Our GNN architecture for coarsening consists of a single graph convolution layer,  
followed by a point-wise multilayer perceptron (MLP) with two fully connected layers and a softmax activation at the end,  
which predicts the soft assignment matrix \( S_{ij} \) for probabilistic clustering.  
As in the referenced work, the features of sources and measurement points remain unchanged during the coarsening process.  
The feature aggregation is mathematically represented as follows:  

\begin{equation}
\mathbf{c}_j = \frac{\sum_{i=1}^{N} S_{ij} \mathbf{x}_i}{\sum_{i=1}^{N} S_{ij}},
\end{equation}
where \( \mathbf{c}_j \) is the centroid feature for cluster \( j \),  
\( \mathbf{x}_i \) is the feature vector for node \( i \),  
and \( S_{ij} \) is the soft assignment of node \( i \) to cluster \( j \).  
Other details on the coarsening algorithm could be found in \cite{Raybov_MLST}.

\begin{figure*}[htb]
    \centering
    \includegraphics[width=0.70\textwidth]{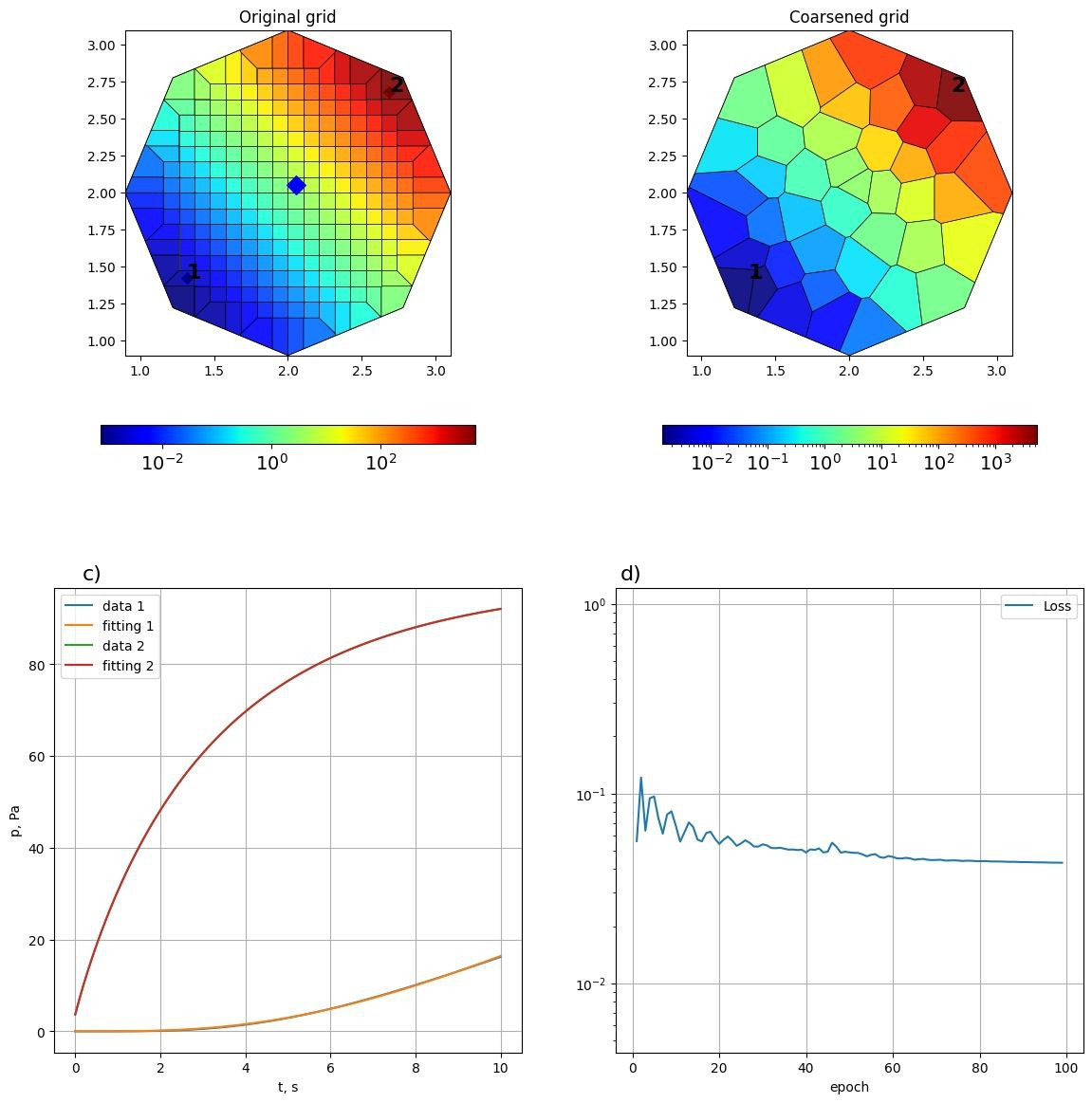}
    \caption{Computational grid coarsening for the linear parabolic equation, \eqref{eq:parab}.
    a) Original grid, color indicates permeability, source marked with blued diamond,
       and two measurement points (1 and 2) marked as black diamonds; 
    b) coarsened grid; 
    c) data at two measurement points, modelled with the original and coarsened grids;
    d) loss function during optimization.}
    \label{fig:ex_1}
\end{figure*}

In the first experiment, the modelling domain was an octagon and partitioned 
into 252 cells, Fig.~\ref{fig:ex_1}~a). The permeability coefficient $K$ varied from $10^{-3}$ to $10^4$ making the problem highly heterogeneous. 
We modelled pressure with \eqref{eq:implicit_scheme} and used the time step, $\tau$, of 0.1 s. 
Pressure values were recorded at two measurement points and
the respective time series are shown in Fig.~\ref{fig:ex_1}~c).
Although the measurement points are symmetric with respect to the source,
they are located in the areas of different permeability, 
leading to different pressure growth rates.

In this experiment, minimization of \eqref{eq:gen_optim} was performed with Adam optimizer and
no stabilizer. 
We picked a coarse grid with 25 cells and tried to optimize it minimizing the misfit between pressure values on the coarse and fine grids over time. 
An approximately optimal grid was received after 200 epochs, Fig.~\ref{fig:ex_1}~b). 

In Fig.~\ref{fig:ex_1}, we observe that the pressure time series match very well and that coarsened grid is fairly regular shaped. This illustrates robustness and stability of our framework.

\subsection{Learnable Graph Coarsening for the Nonlinear Parabolic Equation}
\label{sec:nonliner}

We performed an experiment with learnable graph coarsening
for the nonlinear parabolic equation
governing compressible fluid flow in porous media,
\begin{equation}
\frac{\partial}{\partial t} \left( \phi(u) \rho(u) \right) - \text{div} \left(\rho(u) \frac{K}{\mu}  \nabla u\right) = q,
\label{eq:nonlinear}
\end{equation}
In the nonlinear equation \eqref{eq:nonlinear},
density, $\rho$, and porosity, $\phi$, depend on the pressure,
\begin{equation}
\rho(u) = \rho_r e^{c_{\rho}(u - u_r)}
\end{equation}
\begin{equation}
\phi(u) = \phi_r e^{c_{\phi}(u - u_r)}
\end{equation}
where $\rho_r$, $\phi_r$, $c_\rho$, $c_\phi$ and $u_r$
are some constants. Viscosity $\mu$ is also assumed constant.

To solve \eqref{eq:nonlinear},
we implemented a semi-implicit temporal discretization.
Let us look at the term $\rho(u) \phi(u)$ and define $c(u)$,
\begin{equation}
\frac{\partial}{\partial t} \left( \phi(u) \rho(u) \right) = 
c(u) \frac{\partial u}{\partial t}.    
\end{equation}
Now let us return to \eqref{eq:nonlinear},
\begin{equation}
c(u) \frac{\partial u}{\partial t}
- \text{div} \left(\rho(u) \frac{K}{\mu}  \nabla u\right) = q.
\end{equation}

The semi-implicit temporal discretization takes following form,
\begin{equation}
c(u^n) V \frac{u^{n+1}-u^n}{\tau}
+ A(u^n) \cdot u^{n+1} = V q^{n+1}.
\label{eq:linearized_system}
\end{equation}
Notice the above system is linear although the system matrix 
is updated at every time step.  

\begin{figure*}[htb]
    \centering
    \includegraphics[width=0.70\textwidth]{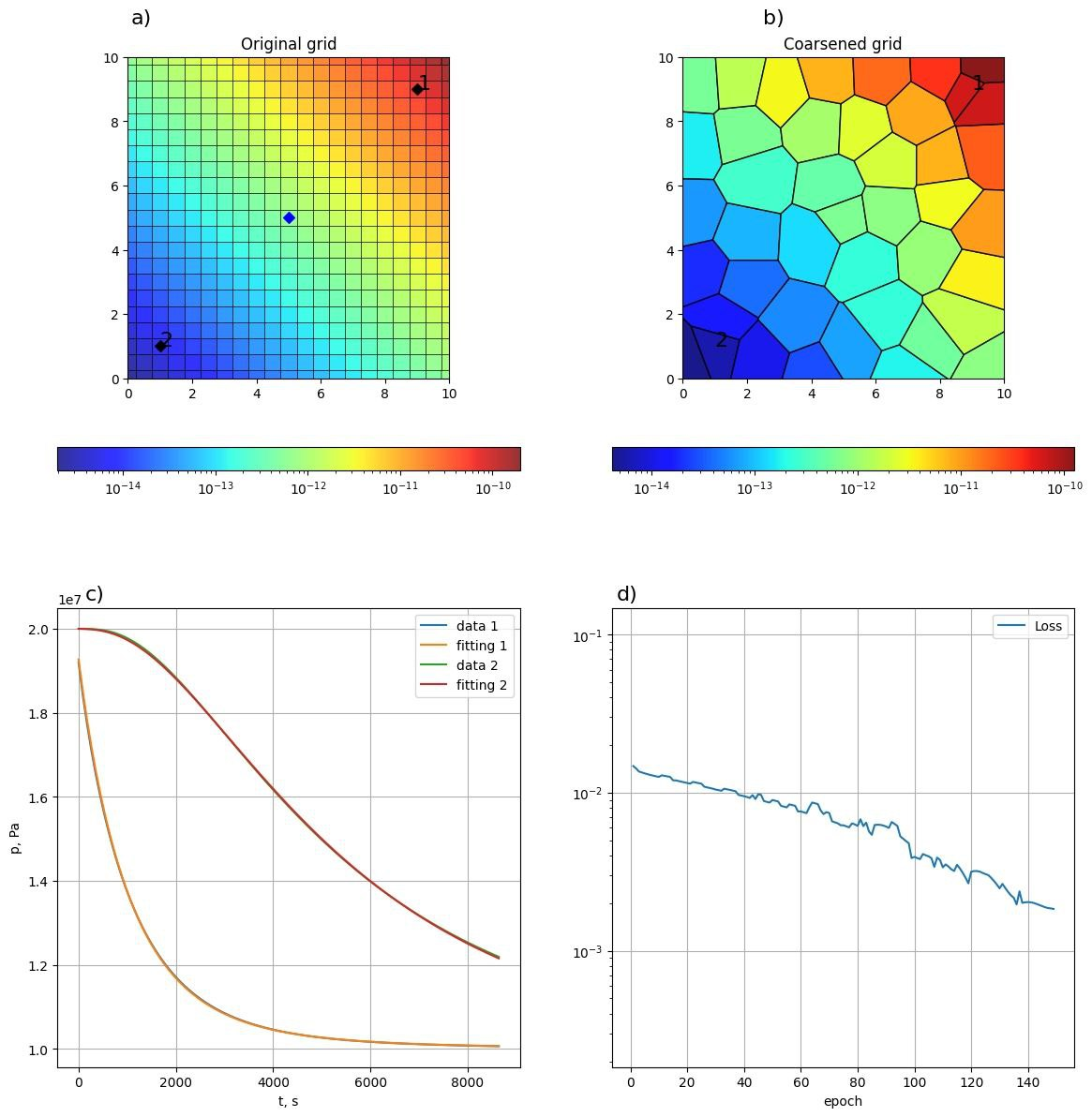}
    \caption{Computational grid coarsening for the nonlinear parabolic equation, \eqref{eq:nonlinear}. 
    a) Original grid, color indicates permeability, source marked with blued diamond,
       and two measurement points (1 and 2) marked as black diamonds; 
    b) coarsened grid; 
    c) data at two measurement points, modelled with the original and coarsened grids;
    d) loss function during optimization.}
    \label{fig:ex_nonl}
\end{figure*}

The experiment was run with next values of parameters: 
Initial pressure $u_0=2\cdot 10^7$ Pa, well pressure in well $u_w=10^7$ Pa, viscosity $\mu=0.005$, $c_{\rho}=10^{-8}$, $\rho_r=850.0$, $c_{\phi}=10^{-11}$, $\phi_r=0.3$ , $u_r=2\cdot 10^7$ Pa. 
Permeability $K$ varied from $10^{-14}$ to $10^{-10}$, Fig.~\ref{fig:ex_nonl}~a).

The results are shown in Fig~\ref{fig:ex_nonl}. We reduced the size of the grid from 400
to 40 cells and obtained a fairly good match of pressure time series.

\subsection{Coefficient Inverse Problem}

Estimation of a poorly known partial differential equation coefficients
describing physical properties of some heterogeneous medium
is of a paramount importance in many applications, especially in geoscience and geophysics. 

We consider a setup similar to the previous experiment.
However, we will now assume that the true permeability,  $K$, is unknown. This problem is known as \textit{history matching} in petroleum engineering, \cite{Srinivasan2022}. We will estimate the permeability by minimizing the following functional, 
\begin{equation}
L(K) =   \int_{0}^T 
\sum_{i=1}^m w_i(t) \left(u_{mod}^i(t) - u_{obs}^i(t) \right)^2 dt 
 + \beta \int_V |\nabla K|^2 dxdy,
\label{eq:inverse_coef}
\end{equation}
constrained with \eqref{eq:parab} and $K(x,y)>0$.
Here, the first term represents the data misfit, while the second term is a stabilizer promoting permeability smoothness. 

\begin{figure*}[htb]
    \centering
    \includegraphics[width=0.60\textwidth]{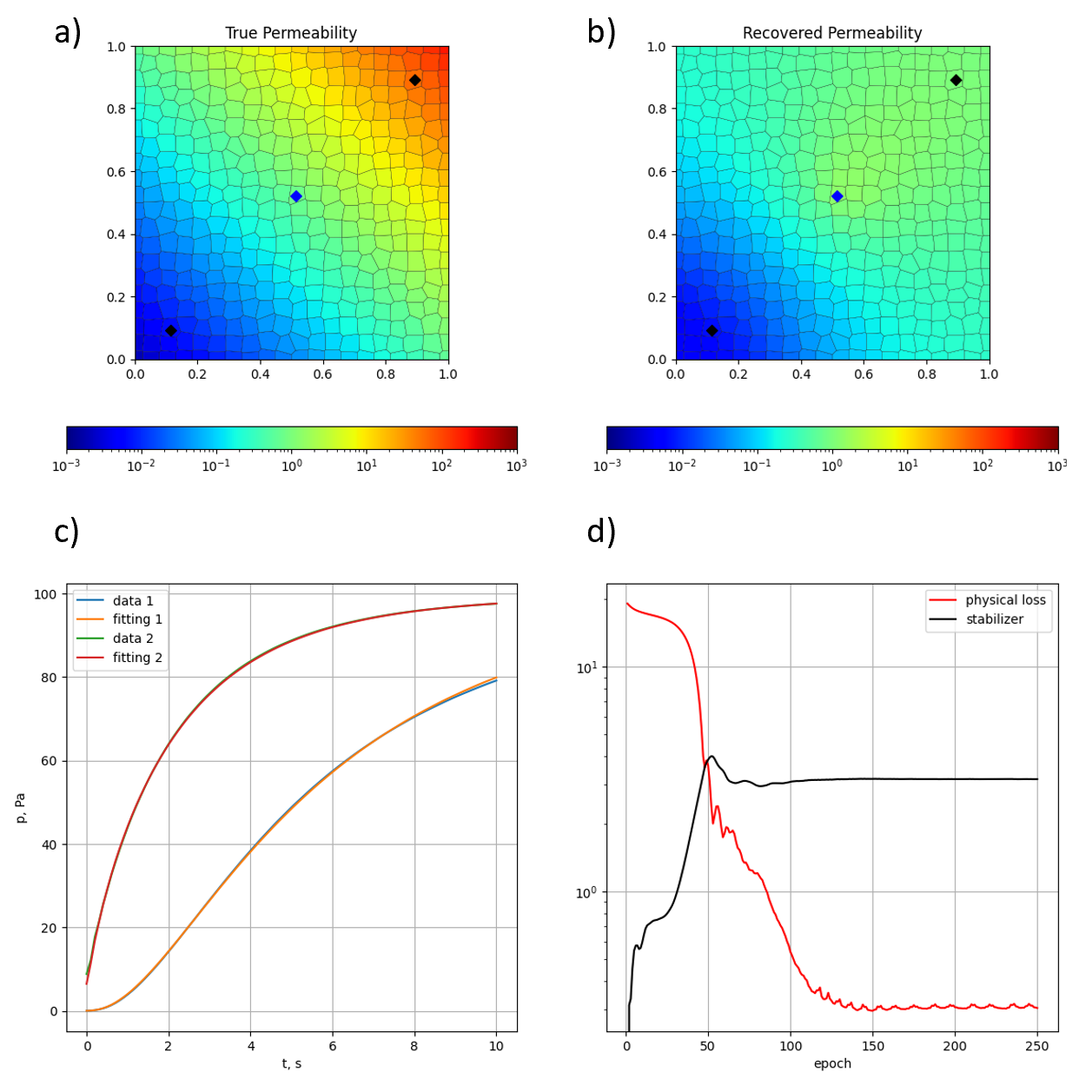}
    \caption{Solution of the coefficient inverse problem. 
    a) True  permeability, source and two measurement points (1 and 2); 
    b) recovered permeability; 
    c) pressure at two measurement points (data and prediction/fitting);
    d) loss function during 250 epochs of minimization.}
    \label{fig:ex_2}
\end{figure*}

In our experiment, pressure was measured at two spatial locations, $m=2$  and
at 100 temporal points, $\tau = 0.1 s$, $T=10 s$, Fig.~\ref{fig:ex_2}~a) and c).
The Voronoi grid was formed from a graph of 400 randomly perturbed vertices.

We applied the Adam optimizer to \eqref{eq:inverse_coef}
stating with $K=1$ and continued for 250 epochs.
Decent data fit was archived after 150 epochs,
Fig.~\ref{fig:ex_2}~d). The recovered permeability,
Fig.~\ref{fig:ex_2}~b), has main features of the true one:
the lower left area of lower  permeability,
the upper right area of higher permeability.
We want to mention that the inverse problems always imply non-uniqueness 
of the solution and this difference between true field and recovered one was expected.

The grid that was used in this experiment was quite distorted, Fig.~\ref{fig:ex_2}a).
We intentionally picked such a grid to demonstrate the flexibility of the presented approach.

\subsection{Inverse Source Problem}

Another type of an inverse problem that is commonly
encountered in geosciences and geophysics is to estimate 
the right-hand side of a partial differential equation.
We will proceed with a setting that is referred to as \textit{self-potential method}, \cite{Revil_Jardani_2013}. The goal is
to estimate direct current source density based on the observations of the electric potential. They are related by the 
equation,
\begin{equation}
    - \textrm{div}(\sigma \nabla u) = f. 
    \label{eq:dc}
\end{equation}

The above equation was considered in a rectangular domain,
completed Dirichlet on the bottom side and Neumann boundary on the other three sides. 
Here $\sigma$ is known electrical conductivity, 
$u$ is the electric potential, $f$ is source density.
Although, the equation is static, its numerical
solution will involve solution of a large sparse linear system,
making it similar to implicit time stepping.
The source density is estimated by minimizing the 
following functional,
\begin{equation}
\begin{split}    
L(f) =  
& 
\sum_{i=1}^n w_i \left(u_{mod}^i - u_{obs}^i \right)^2 \\
+
& \gamma \int_V |\kappa\nabla f|^2 dxdy
+ 
\delta \int_V f^2 dxdy,
\label{eq:inverse_sp}
\end{split}
\end{equation}
where the first term represents data misfit; the second term is a stabilizer promoting smoothness with $\kappa$ being
a variable coefficient needed to compensate decrease of sensitivity with depth;
the third term is a stabilizer promoting sources with smaller energy; $\gamma$ and $\delta$ are trade-off scalar parameters.

\begin{figure*}[htb]
    \centering
    \includegraphics[width=0.85\textwidth]{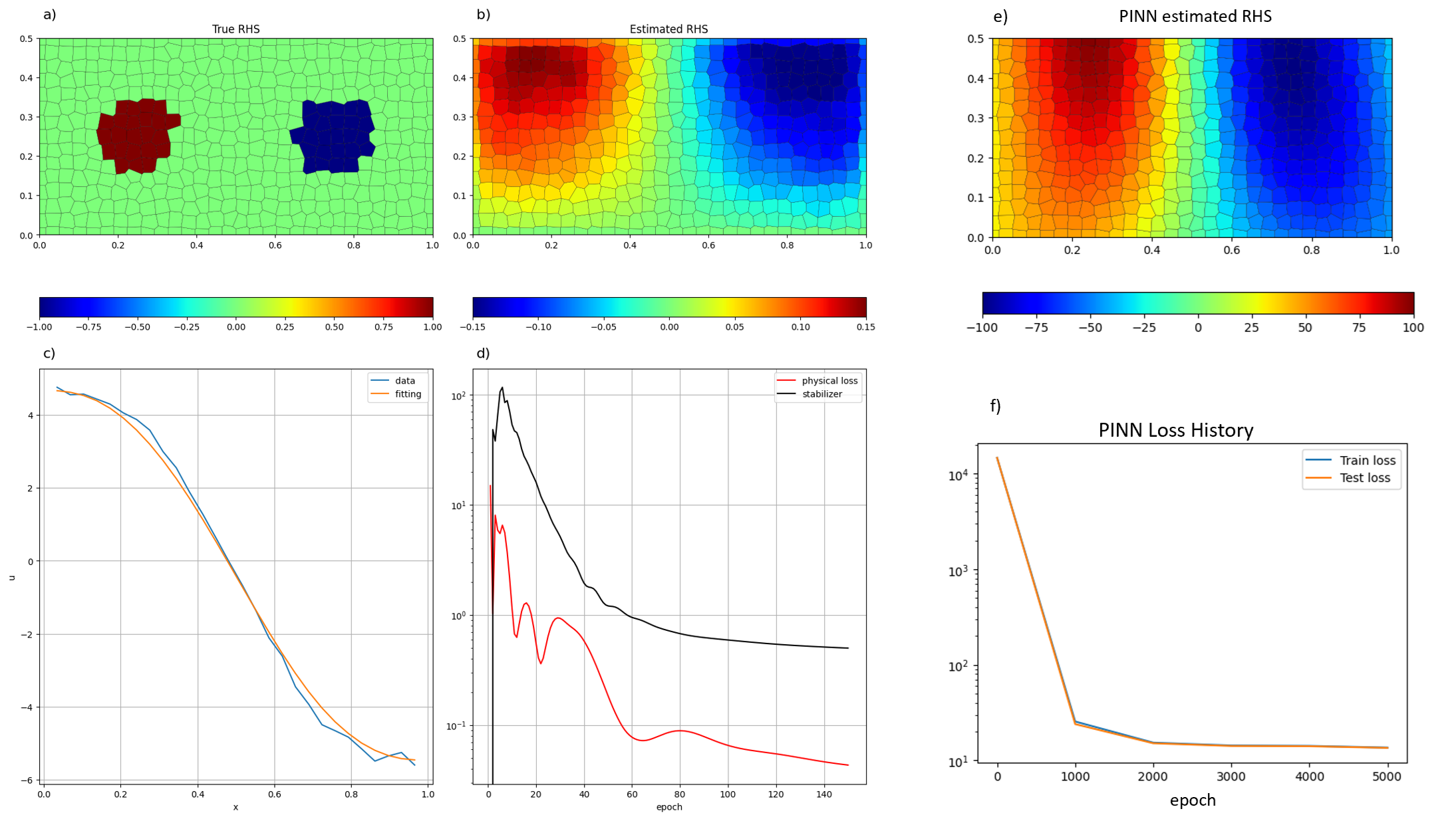}
    \caption{Solution of the source inverse problem. 
    a) True right-hand side $f$ and measurement points (black diamonds); 
      magenta diamonds indicate the side where Dirichlet boundary condition was applied;
      Neumann boundary condition was applied on the other three sides;  
    b) recovered right-hand side; 
    c) observed data at measurement points with with noise added, $u_{obs}$, (data) and prediction, $u_{mod}$, (fitting);
    d) loss function during 100 epochs of minimization: physical loss (first term in \eqref{eq:inverse_sp})
    and stabilizers (later terms in \eqref{eq:inverse_sp}). 
    e) recovered right-hand side $f$ using the PINN;
    f) PINN loss function during 5000 epochs of minimization
    }
    \label{fig:ex_3_rhs}
\end{figure*}

In this experiment, the Voronoi grid was formed from a graph of 900 randomly perturbed vertices.
The electric potential was measured
at $n=28$ points on the upper side of the domain, Fig.~\ref{fig:ex_3_rhs}~a). 
The true right-hand side was formed by a source in the left
part of the domain and a sink in the right part. 
Observed data, $u_{obs}$, was received by solving \eqref{eq:dc} and adding 10\% noise.  
Notice this problem was highly under-determined: we estimated the values of 900 unknowns based on 28. 

We applied the Adam optimizer to \eqref{eq:inverse_sp}
starting with $f=0$ and continuing for 100 epochs.
We observed good data fit and convergence of the optimizer, 
Fig.~\ref{fig:ex_3_rhs}~c) and d). 
The recovered and true right-hand sides,
Fig.~\ref{fig:ex_3_rhs}~b) and ~a), possess common features:
positive values in the left part and negative in the right. The recovered right-hand side is quite smeared. Also the locations of the extrema are biased. 
These artifacts are related to the measurement setup and the
impact of stabilizers. 
The use of more advanced stabilizers can further improve inversion results,
however we also don't expect the full coincidence of the solution to the true sources.

We evaluated our results in this experiment versus those of the PINN
implemented in the DeepXDE package \citep{PINN_2021}. Both the unknown solution $u$ and the right-hand side $f$ were parametrized with a feedforward neural network. The network had two inputs ($x$ and $y$), three dense layers of 150 neurons, and two outputs ($u$ and $f$).
The loss function is defined as the weighted sum of 
$L^2$ norms of residuals for the equation and boundary conditions
and the difference of observed and modelled data.
We had 625 random training residual points in the domain and 900 random points to test the PDE residual.
The network was trained in 5000 epochs
with a learning rate of 1e-3.

Fig.~\ref{fig:ex_3_rhs}~e) shows the estimated right-hand side $f$ using PINN.
We interpolated the values of right-hand side to the same grid as was used above
for easier interpretation. We see a very similar pattern in Fig.~\ref{fig:ex_3_rhs}~b) and Fig.~\ref{fig:ex_3_rhs}~e). 
The main difference between our result and that of PINN 
is in magnitudes. We attribute this to the following.
The measured data received an integral impact of the source and sink.
But since they have same magnitudes and opposite sign, 
the actual magnitudes give a minor impact on the data at measurement points.
Our method captured the magnitudes of the source and sink more correctly than the PINN
due to stabilizers (last two terms in \eqref{eq:inverse_sp}) which were absent in the PINN loss.

Lastly, we note that the convergence of the optimizer in our method was significantly faster, as shown in 
Fig.~\ref{fig:ex_3_rhs}~d) and~f). We attribute this difference to the highly nonlinear nature of the loss functions in PINNs, whereas our approach involves a quadratic loss function, \eqref{eq:inverse_sp}, leading to more efficient optimization.

\section{Conclusions}


In this work, we integrated 
the finite-volume method and implicit time-stepping to develop a fully differentiable modeling pipeline, applying it to both forward and inverse problems of geoscience. Our PyTorch-based pipeline is flexible and easy to implement, enabling the integration of new, fast solvers for large sparse linear systems.

For comparison, \cite{shumilinself} presented results on accelerating forward modeling using graph coarsening, which is somewhat similar to our approach in Section~\ref{ss:fowrd}. However, their method encountered significant time-step limitations due to the use of an explicit scheme.

Both learnable coarsening examples demonstrated that the obtained coarse grid maintains high fidelity to the original simulation while significantly accelerating the modeling process reducing grid size in 10 times (meaning at least 10x speedup).
With implicit time-stepping we easily tackled long-term simulations in highly
heterogeneous media.

%
Our method outperforms the commonly used PINN architecture in solving inverse problems both in terms of quality and computational expenses.

Our pipeline is highly efficient. In future work, we plan to address large-scale 3D problems and more complex equations, such as multi-phase fluid flow. Large-scale 3D modeling and inversion typically rely on custom preconditioned iterative solvers, as they offer both memory efficiency and fast convergence. Our pipeline's ability to easily integrate new solvers makes it particularly appealing for industrial applications.

In this work, we discussed coarsening within the context of forward modeling and data inversion. An interesting future research direction would be to combine coarsening with inversion, allowing data inversion to be performed on an optimal coarse grid in a shorter time.

\section{Acknowledgement}
  The authors are thankful to their colleague Sergei Shumilin for many fruitful discussions.  
  The study was supported by the Russian Science Foundation grant No. 24-41-02035, \url{https://rscf.ru/en/project/24-41-02035/}

\bibliographystyle{plainnat}
\bibliography{my_books}


\end{document}